\documentclass[10pt,twocolumn,letterpaper]{article}

\usepackage{cvm}
\usepackage{times}
\usepackage{epsfig}
\usepackage{graphicx}
\usepackage{amsmath}
\usepackage{amssymb}

\usepackage{subfigure}
\usepackage{caption}
\usepackage{booktabs}
\usepackage{multirow}

\usepackage[pagebackref=true,breaklinks=true,letterpaper=true,colorlinks,bookmarks=false]{hyperref}

\cvmfinalcopy

\def\cvmPaperID{****} 

\ifcvmfinal\fi
\begin{document}

\title{Attention-based~Dual~Supervised~Decoder~for~RGBD~Semantic~Segmentation}

\author{Yang Zhang \thanks{Work done while interning at Hubei University of Technology}\\
Nanjing University\\
{\small \url{https://yangzhangcst.github.io/Homepage/}}
\and
Yang Yang  \\
Nanjing University\\
{\tt\small yyang\_nju@outlook.com}
\and
Chenyun Xiong\\
Hubei University of Technology\\
{\tt\small cyx@hbut.edu.cn}
\and
Guodong Sun\thanks{Corresponding author}\\
Hubei University of Technology\\
{\tt\small sgdeagle@163.com}
\and
Yanwen Guo \thanks{Corresponding author}\\
Nanjing University\\
{\tt\small ywguo@nju.edu.cn}
}

\maketitle

\begin{abstract}
	Encoder–decoder models have been widely used in RGBD semantic segmentation, and most of them are designed via a two-stream network. In general, jointly rea-soning the color and geometric information from RGBD is beneficial for semantic segmentation. However, most existing approaches fail to comprehensively utilize multi-modal information in both the encoder and decoder. In this paper, we propose a novel attention-based dual supervised decoder for RGBD semantic segmentation. In the encoder, we design a simple yet effective attention-based multi-modal fusion module to extract and fuse deeply multi-level paired complementary information. To learn more robust deep representations and rich multi-modal information, we introduce a dual-branch decoder to effectively leverage the correlations and complementary cues of different tasks. Extensive experiments on NYUDv2 and SUN-RGBD datasets demonstrate that our method achieves superior performance against the state-of-the-art methods.
\end{abstract}

\section{Introduction}\label{sec:introduction}

In recent years, scene understanding has received considerable attention due to the wide applications in AR/VR~\cite{Ping:VR19}, autonomous driving~\cite{Brickwedde:ICCV19,Xu:ICCV19}, UAVs~\cite{Teixeira:RA20}, simultaneous localization and mapping (SLAM)~\cite{Ma:ICRA16}, Robotics~\cite{Marchal:RA20}, and other artificial intelligence fields. As a result, semantic segmentation for scene understanding becomes extremely important. However, there still exists many challenges in RGBD semantic segmentation caused by the complexity of the environment, the influence of inaccurate depth, and the joint reasoning of multi-modal information.

\begin{figure}[!t]
	\centering
		\subfigure[Atrous Conv.]{
			\includegraphics[width=0.8in]{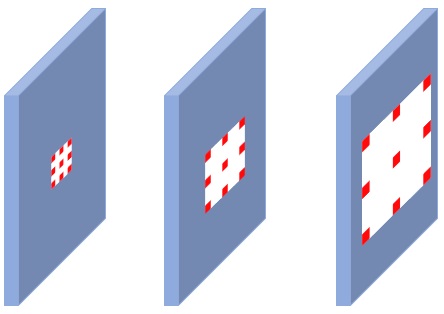}
		}\quad
		\subfigure[Encoder-decoder for segmentation]{
			\includegraphics[width=2.0in]{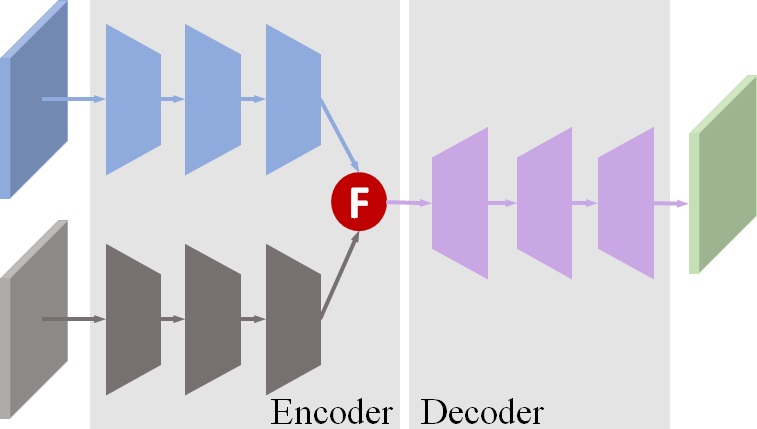}
		}
		
		\subfigure[Encoder-decoder for multi-task including segmentation]{
			\includegraphics[width=2.9in]{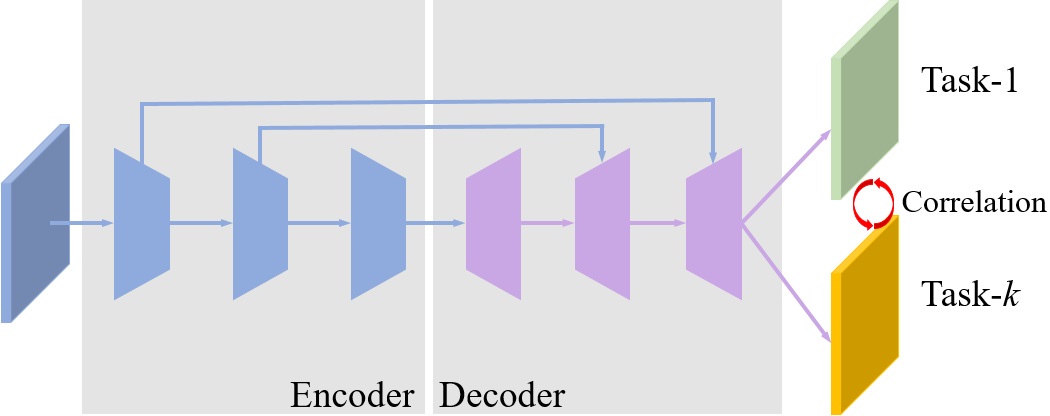}
		}
		
		\subfigure[Dual supervised decoder for segmentation (ours)]{
			\includegraphics[width=2.9in]{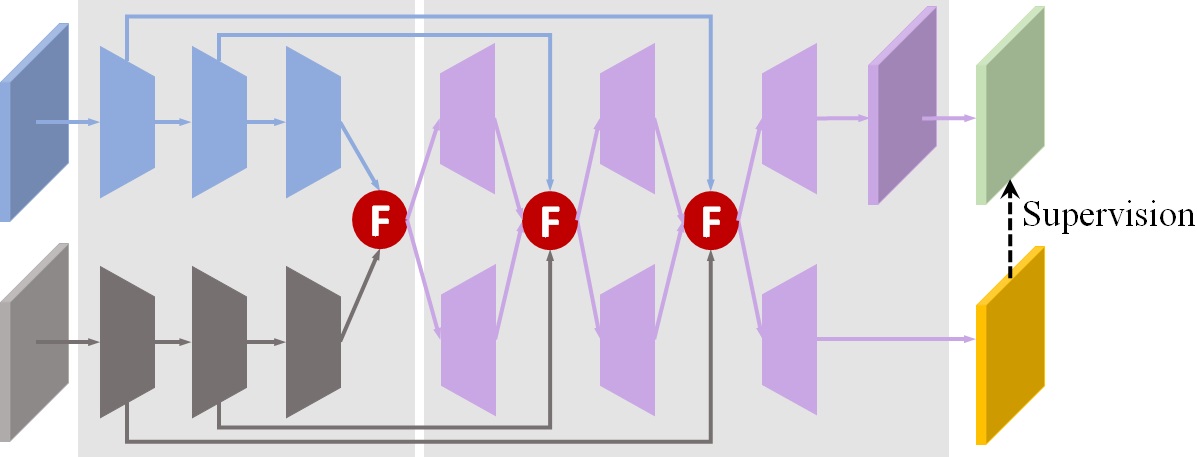}
		}
	\caption{Examples of typical structures for RGBD semantic segmentation. The blue color and gray color indicate the RGB and depth streams, separately. The \textcircled{F} denotes the combination operation. }
	\label{fig:intro}
\end{figure}

Deep learning technique has been applied to the semantic segmentation problem with great success. Though different architectures are developed, the convolutional neural networks (CNNs) are still prevalent due to their ability to model non-linear, high-dimensional functions.  Generally, atrous/dilated convolution-based methods~\cite{Chen:ICL15,Gadde:ECCV16,Lin:ICCV17,Lin:TPAMI20} allow us to effectively enlarge the field-of-view of filters to incorporate multi-scale context (see Fig.~\ref{fig:intro}(a)), especially for atrous spatial pyramid pooling (ASPP)~\cite{Chen:TPAMI18}. But there exists a `gridding' problem~\cite{Wang:WACV18}, and they fail to capture small objects with accurate boundaries. Furthermore, it is computationally intensive if denser output features are extracted for this type of models.

The encoder-decoder models~\cite{Noh:ICCV15,Badri:TPAMI17, Kendall:BMVC17,Lee:ICCV17,Tang:arXiv20} allow for faster computation in the encoder path and recovering sharp object boundaries in the decoder. These models, however, only use RGB data for semantic segmentation which cannot achieve a satisfactory performance.
Compared with color, the depth data provide geometric cues to reduce the uncertainty of the segmentation of objects in which the color is similar to the background~\cite{Hazirbas:ACCV17}. It is thus meaningful and crucial to develop effective models to combine these complementary modalities for segmentation. To achieve this goal, numerous works~\cite{Hazirbas:ACCV17,Cheng:CVPR17,Jiang:arXiv18,Hu:ICIP19,Chen:arXiv20,Xing:ECCV20,Chen:ECCV20} focus on designing a two-stream network which processes the RGB and geometry information in terms of depth or HHA, separately. As shown in Fig.~\ref{fig:intro}(b), the features from two modalities are further fused by various mechanisms such as the element-wise summation~\cite{Hazirbas:ACCV17,Jiang:arXiv18}, gate~\cite{Cheng:CVPR17,Chen:ECCV20}, and attention~\cite{Hu:ICIP19,Seichter:arXiv20} in the encoder. Such approaches only process the paired complementary cues in the encoder, but ignoring the cross-modal information during decoding. Moreover, training such a model is usually difficult to converge due to this imbalance of the encoder and decoder.
\\
Since other related tasks such as depth estimation could facilitate semantic segmentation, recent works~\cite{Eigen:ICCV15,Kong:CVPR18,Zhang:ECCV18,Nekrasov:ICRA19,Zhang:CVPR19,Zhou:CVPR20} have attempted to solve the segmentation problem via a multi-task learning framework. Fully convolutional encoder-decoder networks have become the mainstream. During the joint learning, different task-specific decoders explore the correlations between these tasks as shown in Fig.~\ref{fig:intro}(c). Note that these methods perform the multi-task distillation at a fixed scale (\textit{i.e.} backbone features) with specific receptive field in the decoder. However, in fact, the influence between two tasks is different for various sizes of receptive field~\cite{Vandenhende:ECCV20}. Furthermore, the capacity of fully convolutional encoder-decoder, whose encoder and decoder are simply integrated together (\emph{e.g}. skip connection~\cite{Zhang:ECCV18,Nekrasov:ICRA19,Zhou:CVPR20}, multi-scale feature aggregation~\cite{Zhang:CVPR19}), is limited for such a complex task of semantic segmentation.

In this paper, we design a simple symmetric yet effective network (in Fig.~\ref{fig:intro}(d)) to efficiently use the multi-level cross-modal information for RGBD semantic segmentation. Motivated by the above observations, we first propose an attention-based multi-modal fusion module to process the multi-level paired complementary information in a two-stream encoder. To learn cross-modal information during decoding, we introduce a novel dual-branch decoder in which the primary is designed for semantic segmentation supervised by another task-guided branch. Such design enables us to incorporate multi-scale context by the ASPP module at the end of primary-branch, which contains the pyramid supervision for enhancing the deep representation. This specific dual-branch decoder is capable of improving the performance of semantic segmentation through multi-task distillation, while facilitating the convergence of training to solve the imbalance problem of the encoder and decoder. We conduct experiments on the NYUDv2 and SUN-RGBD datasets to validate the superior performance of our method in comparison with the state-of-the-arts.

Our contributions are summarized as follows.

\begin{itemize}
	\item{We propose a novel attention-based dual supervised decoder to utilize the complementary information across modalities for RGBD semantic segmentation.}
	
	\item{ We design a simple yet effective attention multi-modal fusion module to extract and fuse deeply multi-level paired complementary information.}
	
	\item{ We propose a dual-branch decoder to learn more robust deep representations and rich multi-modal information for the improvement of semantic segmentation performance and the efficiency of training. }
	
	\item{ The proposed method achieves superior performance against the state-of-the-art methods on public benchmark datasets.}
\end{itemize}

\begin{figure*}[!t]
	\centering
	\includegraphics[width=6.5in]{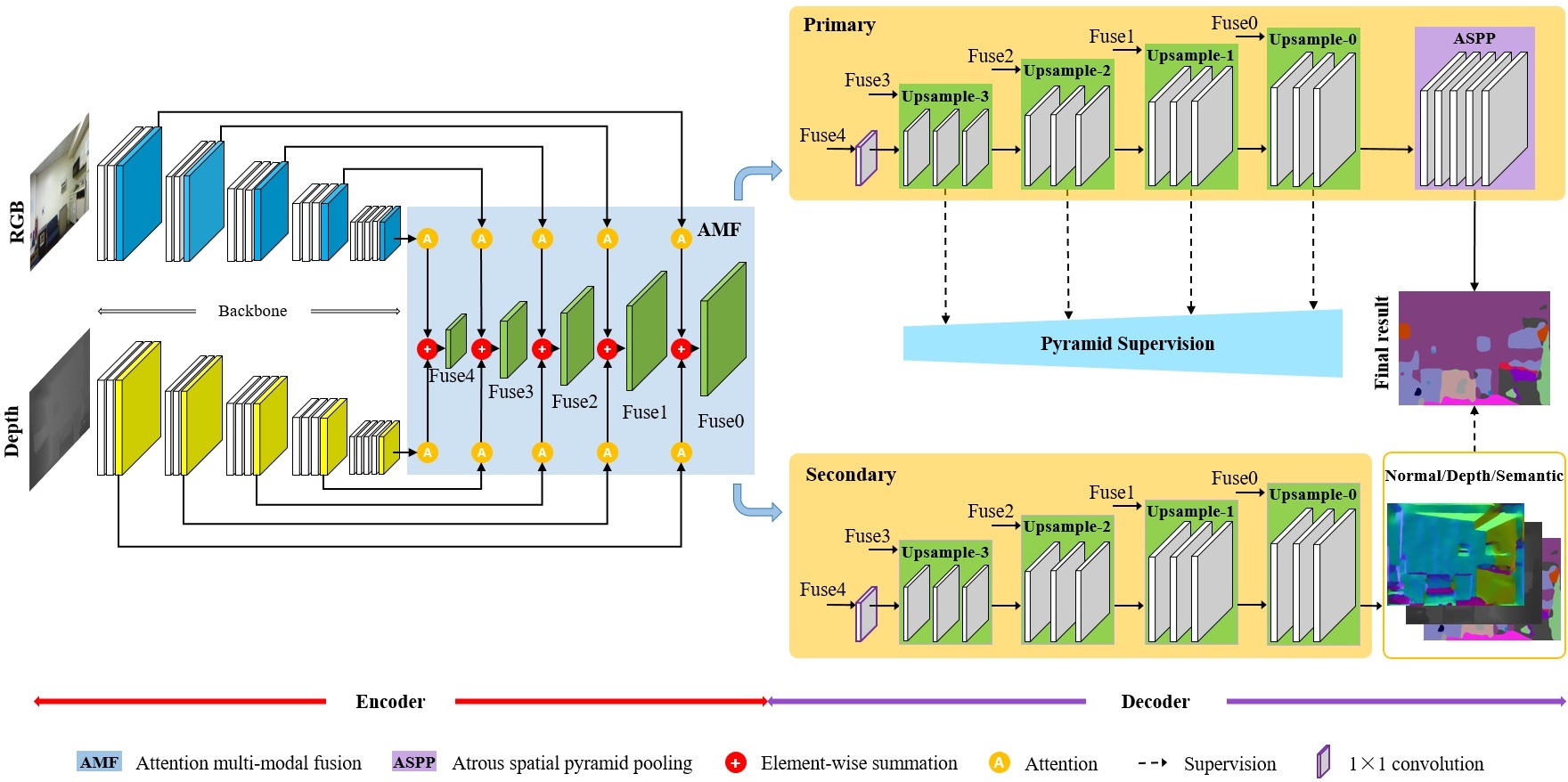}
	\caption{Overview of the proposed ADSD architecture. We employ a two-stream encoder and a dual-branch decoder. The input of the network is a pair of RGB-Depth images. The feature maps of backbone encoders are fused through AMF module, which are further used to output the results through upsampling modules in the dual-branch decoder. At the end of primary branch, the ASPP is introduced to improve the final segmentation performance. Meanwhile, each upsampling block predicts a side output for pyramid supervision. In addition to the semantic supervision, the secondary branch requires supervision from normal estimation, depth estimation, or semantic segmentation task.}
	\label{Fig:framework}
\end{figure*}

\section{Related work} \label{sec:Relatedwork}

In recent years, CNN-based methods have been successfully applied to the RGBD semantic segmentation\footnote{https://github.com/Yangzhangcst/RGBD-semantic-segmentation}. In terms of structure, these methods can be roughly divided into the following three groups.

\textbf{Atrous/dilated Convolution.}
Several works~\cite{Chen:ICL15,Gadde:ECCV16,Lin:ICCV17,Qi:ICCV17,Wang:ECCV18,Lin:TPAMI20} utilized the atrous/dilated convolution to incorporate multi-scale context for RGBD semantic segmentation. For example, Chen \textit{et al.}~\cite{Chen:ICL15} proposed a dilated convolution which can enhance the receptive field while keep the resolution of the feature map. Qi \textit{et al.}~\cite{Qi:ICCV17} introduced a 3D graph neural network (3DGNN) to model accurate context with geometry cues provided by depth based on the dilated convolution. Lin \textit{et al.}~\cite{Lin:TPAMI20} presented RefineNet, a generic multi-path refinement network that explicitly exploits all the information available along the down-sampling process to enable high-resolution prediction using long-range residual connections. However, dilated convolution can result in losing the continuity of feature maps. In addition, it is only effective for some large objects and invalid for small objects, which is not helpful to extract accurate edges.

\textbf{Encoder-decoder.}
Many efforts~\cite{Noh:ICCV15,Badri:TPAMI17,Cheng:CVPR17,Hazirbas:ACCV17,Kendall:BMVC17,
	Shelhamer:TPAMI17,Lee:ICCV17,Jiang:arXiv18,Tang:arXiv20,Zhou:ACCV20,Chen:arXiv20,Xing:ECCV20,
	Chen:ECCV20,Seichter:arXiv20} concerning encoder-decoder architectures have been devoted to RGBD semantic segmentation. For instance, DeconvNet~\cite{Noh:ICCV15} used stacked deconvolutional layers to produce high-resolution prediction and more semantic details. SegNet~\cite{Badri:TPAMI17} shared a similar idea using indices in pooling layers to promote the recovery process. To learn the optimal fusion of multi-modal features, RDFNet~\cite{Lee:ICCV17} extended the core idea of residual learning to RGBD semantic segmentation. Hu \textit{et al.}~\cite{Hu:ICIP19} proposed a architecture ACNet with three parallel branches and a channel attention-based module that extracts weighted features from RGB and depth branches.  Chen \textit{et al.}~\cite{Chen:arXiv20} proposed a spatial information guided convolution network (SGNet) which allows to integrate 2D and 3D spatial information. ESANet~\cite{Seichter:arXiv20} used two ResNet-based encoders with an attention-based fusion for incorporating depth information, and a decoder utilizing a learned upsampling. However, these methods only perform the multi-modal information in the encoder, but ignore the cross-modal cues in the decoder. Moreover, when a large number of encoder parameters are passed to the decoder, it is difficult to train such a model to converge quickly.

\textbf{Multi-task Learning.}
Numerous works ~\cite{Eigen:ICCV15,Kong:CVPR18,Xu:CVPR18,Zhang:ECCV18,Nekrasov:ICRA19,Zhang:CVPR19,Zhou:CVPR20,Vandenhende:ECCV20} have also explored the idea of combining networks for complementary tasks to improve learning efficiency and generalization across different tasks. For example, Eigen \textit{et al.}~\cite{Eigen:ICCV15} proposed a single multi-scale network (MSCNN) to address three different computer vision tasks. Zhang \textit{et al.}~\cite{Zhang:ECCV18} proposed a joint task-recursive learning (TRL) framework to refine the results of both semantic segmentation and monocular depth estimation through serialized task-level interactions. Zhang \textit{et al.}~\cite{Zhang:CVPR19} proposed a pattern affinitive propagation (PAP) method to utilize the matched affinity information across tasks.  Zhou \textit{et al.}~\cite{Zhou:CVPR20} proposed intra-task and inter-task pattern-structure diffusion (PSD) to learn long-distance propagation and transfer cross-task structures. Different from the previous works, we incorporate multi-modal information in the both encoder and decoder through attention-based dual supervised decoder to provide a unified pixel-wise scene understanding.

\section{Method}
In this section, we describe the proposed attention-based dual supervised decoder (ADSD) in detail. First, we briefly describe the overall architecture. Then, we discuss multi-level fusion strategy and attention block used in attention-based fusion module for multi-modal features in the encoder. Moreover, we give a detailed depiction of our dual-branch decoder which significantly improves the performance of semantic segmentation. Finally, we introduce the objective function for optimizing the network.

\subsection{The Network Architecture}\label{sec:Method}
The entire network architecture of our ADSD is presented in Fig.~\ref{Fig:framework}. For clear illustration, we use blocks with different colors to indicate different layers. Note that each convolution layer in our network is followed by a batch normalization layer~\cite{Ioffe:ICML2015} before the activated function of rectified linear unit (ReLU), and it is omitted in the figure for simplification. The whole network can be divided into a two-stream encoder and a dual-branch decoder. In the decoder, the primary branch with pyramid supervision is designed for semantic segmentation, and the secondary branch requires supervision from the other task such as normal estimation, depth estimation, or semantic segmentation.

In the encoder part, we design two independent branches to extract features from RGB and depth images separately. In these two branches, we simply choose ResNet-50~\cite{He:CVPR2016} as the backbone to extract multi-scale hierarchical feature maps from inputs. The output features from RGB and depth branch are combined to produce fusion features (Fuse0$\sim$Fuse4) through the attention-based multi-modal fusion (AMF) module, where the details are given in Section~\ref{sec:encoder}. It is worth noting that there is no connection between fusion features at different scales.

In the decoder part, we feed the above fusion features into each task-branch to decode pixel-level information. To produce high resolution predictions, we decode these convolutional features and then combine with the same scale fused features by upsampling blocks to produce task-specific features as shown in Section~\ref{sec:decoder}. Specially, at the end of the primary branch, the ASPP is introduced to improve the final segmentation performance. Meanwhile, each upsampling block predicts a side output for pyramid supervision, which are introduced in Section~\ref{sec:supervision}. 
\begin{figure}[t]
	\centering
		\subfigure[Channel attention]{
			\includegraphics[width=3in]{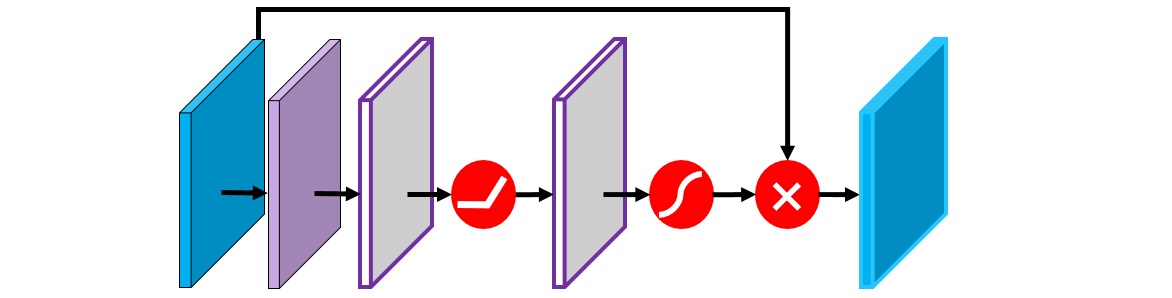}
		}
		
		\subfigure[Spatial attention]{
			\includegraphics[width=3in]{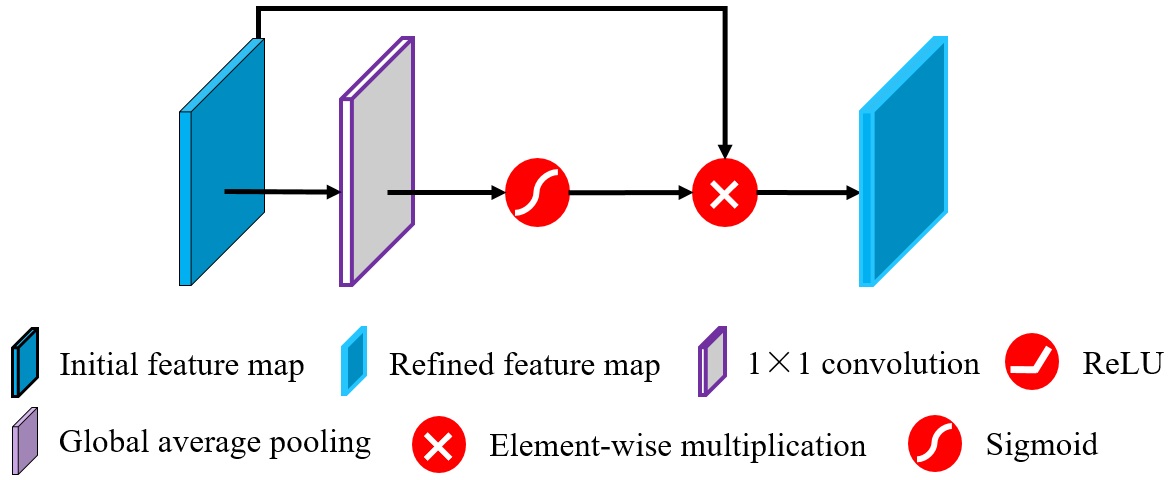}
		}
	\caption{Detailed structure of channel attention and spatial attention.}
	\label{Fig:Attention}
\end{figure}

\subsection{Encoder}\label{sec:encoder}
The conventional fusion branch~\cite{Hu:ICIP19,Tang:arXiv20} integrates multi-scale features by coarse-to-fine CNNs and general attention mechanisms. Such approaches are computationally expensive leading to information redundancy easily. Considering the complementarity between paired RGB and depth cues in multiple layers, we design a simple yet effective AMF module to fully extract and fuse multi-level paired complementary information. As illustrated in the middle part of Fig.~\ref{Fig:framework}, we show the main process of AMF while leveraging the high performance of the attention block. In our implementation, our AMF includes all five scales (\emph{i.e.}1/2, 1/4, 1/8, 1/16, 1/32) of the backbone network.

To improve the performance of semantic segmentation, the channel attention~\cite{ZhangYL:ECCV18} allows the network to concentrate on more useful channels and flattens the distribution of information among channels with the effective utilization of complementary features. The architecture of the channel attention is illustrated in Fig.~\ref{Fig:Attention}(a). Assuming an input feature map $U=[u_1, u_2,...,u_C]\in\mathbb{R}^{C\times H\times W}$ that passes through channel attention block $A_C(\cdot)$ to generate output feature map $V_{ac}\in\mathbb{R}^{C\times H\times W}$. Here, $H$ and $W$ are the height and width respectively, with $C$ being the number of channels. Channel attention is firstly performed by a global average pooling to produce a vector $Z\in\mathbb{R}^{C\times 1\times 1}$ with its $t$-th element
\begin{equation}
	\label{eq1}
	Z_t=\frac{1}{H\times W}\sum_{i}^{H}\sum_{j}^{W}u_t(i,j)\;.
\end{equation}
Then $Z$ is transformed to $\hat{Z}=W_{1\times1}(\delta (W_{1\times1}(Z)))$, with $W_{1\times1}$ being the weight of a 1$\times$1 convolutional layer and the ReLU operator $\delta(\cdot)$. A sigmoid $\sigma(\hat{Z})$ is applied to activate the convolution result, constraining the value of weight vector to the interval [0,1]. Finally, we perform an element-wise multiplication, and the result $V_{ac}$ can be expressed as:
\begin{equation}
	\label{eq2}
	V_{ac}=A_C(U)=[\sigma(\hat{Z_1})u_1,\sigma(\hat{Z_2})u_2,...,\sigma(\hat{Z_C})u_C]\;
\end{equation}

In contrast to channel attention, spatial attention~\cite{Roy:MICCAI18,Hu:TPAMI20} has fewer parameters with a simpler structure. The architecture of the spatial attention is illustrated in Fig.~\ref{Fig:Attention}(b). We consider an alternative slicing of an input tensor $U=[u^{1,1},...,u^{i,j},...,u^{H,W}]$ that passes through the spatial attention block $A_S(\cdot)$ to generate output $V_{sc}$, where $u^{i,j}\in\mathbb{R}^{C\times 1\times 1}$ corresponding to the spatial location $(i,j)$. The spatial attention is firstly performed by a 1$\times$1 convolution to generate a projection tensor $Q\in\mathbb{R}^{H\times W}$. Each $Q_{i,j}$ of the projection describes the linearly combined representation of a spatial location $(i,j)$. This projection is then performed on a sigmoid $\sigma(\cdot)$ to rescale activations to [0,1]. And the result $V_{sc}$ can be expressed as
\begin{equation}
	\label{eq3}
	\begin{split}
		V_{sc}=A_S(U)=[&\sigma(Q_{1,1})u^{1,1},\sigma(Q_{1,2})u^{1,2},...,\\
		&\sigma(Q_{i,j})u^{i,j},...,\sigma(Q_{H,W})u^{H,W}]\;.
	\end{split}
\end{equation}
Specifically, this operation provides more importance to relevant spatial locations and ignores irrelevant ones.

\subsection{Decoder}\label{sec:decoder}
\begin{figure}[t]
	\centering
	\includegraphics[width=3.1in]{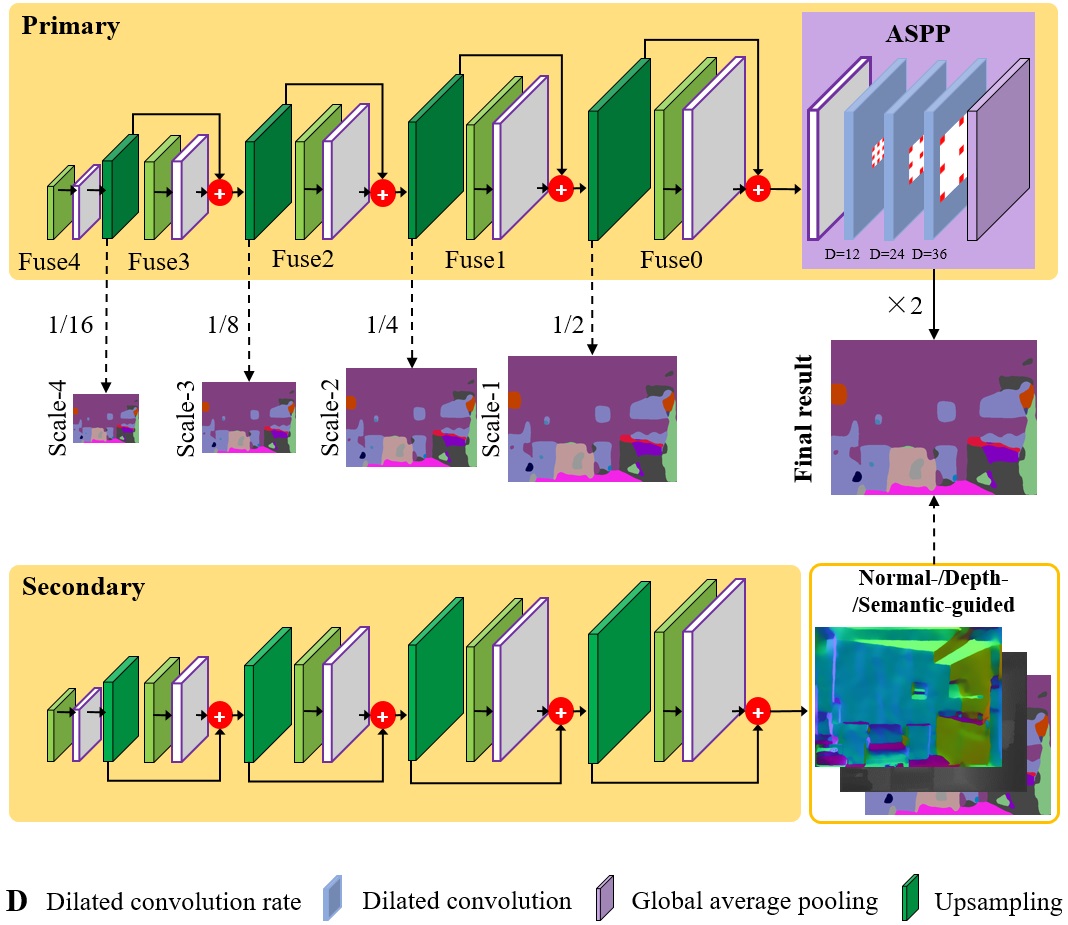}
	\caption{Detailed diagram of the proposed dual-branch decoder. The primary branch computes multi-scale fused features through 1$\times$1 convolutional layers and upsampling blocks. The final features is refined by the ASPP to predict segmentation result, which is supervised via the output generated by normal-/depth-/semantic-guided branch.}
	\label{Fig:decoder1}
\end{figure}

Benefit from the exploration of correlation between different tasks in multi-task learning~\cite{Xu:CVPR18,Zhang:CVPR19, Zhou:CVPR20}, we propose a novel dual-branch decoder to learn more robust deep representations and multi-modal information. It is well-known that low-level layers of the CNNs usually have more positional information, while high-level layers contain more semantic cues. Both the positional and semantic cues play a key role in semantic segmentation. Inspired by upsampling strategy in~\cite{Hu:ICIP19} and skip connection like~\cite{He:CVPR2016}, we use transposed convolutional layers to upsample the features at different pyramid scales, as illustrated in Fig.~\ref{Fig:decoder1}.

In particular, the fused feature map $V'_K$ of AMF is firstly calculated by a 1$\times$1 convolution $W_{1\times1}$ to project the feature map $W_{1\times1}(V'_K)$ with lower channel, allowing the decoder to have a lower memory consumption. And it passes through upsampling block $B_U(\cdot)$ to generate the feature map $S_{K}$ of $K$-th slide output.
\begin{equation}
	\label{eq4}
	S_{K}=B_U[W_{1\times1}(V'_K)]\;.
\end{equation}
\label{eq5}
Then $S_{K}$ is used to produce the next slide output as follows:
\begin{equation}
	S_{K-1}=S_{K}{\textcircled{\scriptsize{+}}}B_U[W_{1\times1}(V'_{K-1})]\;,
\end{equation}
where {\textcircled{\scriptsize{+}}} is element-wise summation. Repeatedly, we continue to upscale feature maps and perform the above decoding process to produce a higher scale of feature maps. The scale factor of each upsampling block is set to 2. All slide outputs are employed for pyramid supervision which will be introduced in Section~\ref{sec:supervision}. In particular, the ASPP is introduced to incorporate multi-scale context at the end of this branch. In our experiments, the dilated convolution rate is set as 12, 24, and 36 in the ASPP.

In secondary branch of decoder, we repeat the operations on the primary branch to upsample the fused feature map $V'_a$. The final upsampling feature map $S_{0}$ is directly used to generate the predict which can be surface normal, estimated depth, or segmentation result. In practice, we propose a more efficient training method that takes advantage of multi-modal feature sharing during training. Inspired by the training strategy in~\cite{Girshick:CVPR2014}, we train the model for a depth-guided branch decoder at the pre-training stage and the semantic-guided branch decoder at the fine-tuning stage.

\subsection{Objective Function}\label{sec:supervision}
\noindent
\textbf{Pyramid Supervision.}
The pyramid supervised training scheme alleviates the gradient disappearance problem by introducing supervised learning at different levels~\cite{Jiang:arXiv18}. As shown in Fig.~\ref{Fig:decoder1}, the primary-branch of decoder computes $K$ slide outputs by upsampling blocks with different spatial resolutions. In our implementation, the $K$ is set to 4, and the slide outputs are defined as Scale 1 to 4. The resolution scales are 1/2, 1/4, 1/8, and 1/16, and the final result is a full resolution. We calculate the score map of each output through a 1$\times$1 convolution, and then feed it into a softmax layer and cross-entropy function to build the loss function $L_{P_k}$ ($k\in[1,K]$).

\noindent
\textbf{Loss Function.}
For semantic segmentation, most methods utilize cross-entropy to measure the difference between the prediction and ground-truth. However, for existing datasets, the distribution of semantic labels is extremely imbalanced. This will bias the learning towards the dominant samples and lead to low accuracy in minority categories. To alleviate the data imbalance issues, we re-weight the training loss of each class in the cross-entropy function using the median frequency setting proposed in~\cite{Eigen:ICCV15,Jiang:arXiv18}. That is, we weight each pixel by a factor of $\alpha_c=p_m /p_c$, where $c$ denotes the ground-truth category. $p_c$ is the pixel probability of that category, $p_m$ is the median of all the probabilities of these categories.

For different task supervision, we use task-guided loss functions defined as $L_{T}$ which can be normal $L_{N}$, depth $L_{D}$ or semantic $L_{S}$. Following the depth estimation algorithms, we use berHu loss~\cite{Laina:3DV16} for the depth supervision:
\begin{equation}
	\label{eq6}
	L_D=\sum_{i}
	\begin{cases}
		|d_i-D_i|, &  |d_i-D_i|\leqslant \beta\\
		\frac{(d_i-D_i)^2+\beta^2}{2\beta}, & |d_i-D_i|> \beta
	\end{cases},
\end{equation}
where $d_i$ is the predicted depth for pixel $i$, and $D_i$ is the ground-truth. $\beta=\frac{1}{5}max(|d_i-D_i|)$. Such a loss function can provide more obvious gradients at the locations where the depth difference is low, and thus can help to better train the network. As for surface normal, we also use the berHu loss~\cite{Laina:3DV16}. Together with the above pyramid supervision loss $L_{P_k}$ for semantic prediction at intermediate layers, the total loss $L$ can be defined as:
\begin{equation}
	\label{eq7}
	L=L_S+L_{T}+\sum_{k=1}^{K}L_{P_k}\;.
\end{equation}
Finally, a fully end-to-end optimization is computed by using gradient back-propagation.

\section{Experiments}\label{sec:Experiments}
To evaluate our proposed method, we conduct extensive experiments on NYUDv2 dataset~\cite{Silberman:ECCV12} and SUN-RGBD dataset~\cite{Song:CVPR15}. We start with the introduction of experimental setup such as implementation details, datasets, and evaluation metrics. We then conduct ablation experiments to determine whether our network improve performance. Finally, we compare our method with the existing methods for semantic segmentation on these datasets.

\subsection{Implementation Details}\label{sec:details}
We implement our method using the publicly available Pytorch. For the optimizer, we use Adam~\cite{Kingma:arXiv14} with $(\beta_1,\beta_2)=(0.9,0.999)$. For NYUDv2 dataset, we train the model for 600 epochs and fine-tune 50 epochs with a learning rate of 0.0002 and 0.00002, respectively. For SUN-RGBD dataset, we train the model for 300 epochs and fine-tune it for 30 epochs with the same learning rate. We adopt the step learning rate policy whose learning rate is updated after each 300 epochs. Specifically, all experiments are trained with batch size 8 on a single NVIDIA Tesla V100 GPU. To avoid overfitting, similarly with~\cite{Chen:ECCV20,Ding:TIP20,Lin:TPAMI20}, we employ general data augmentation strategies, including random scaling in the range of [0.8, 1.4], random horizontal flipping, and random cropping. In particular, we resized the inputs to a resolution of 480$\times$640 for the above datasets. During the inference, we only obtain the prediction results from the primary decoder for semantic segmentation.

\subsection{Datasets and Metrics}
We use the NYUDv2 dataset~\cite{Silberman:ECCV12} for the main evaluation of our method and further use the SUN-RGBD dataset~\cite{Song:CVPR15} for extensive comparison with the state-of-the-arts. The NYUDv2 dataset consists of 1449 RGBD images showing interior scenes. We use the segmentation labels provided in~\cite{Saurabh:CVPR13}, in which all labels are mapped to 40 classes. We use the standard training/test split with 795 and 654 images, respectively. The SUN-RGBD contains 10335 RGBD images labeled with 37 classes. We use the official training set with 5285 images to train our network, and the official testing set with 5050 images for evaluation. Compared with NYUDv2, SUN-RGBD has more complex scene and depth conditions, which are probably more suitable to measure the generality of our method. For the evaluation of semantic segmentation results, we follow the recent works~\cite{Chen:ECCV20,Ding:TIP20,Lin:TPAMI20,Zhang:CVPR19,Zhou:CVPR20} and use three common metrics for evaluation, including pixel accuracy (PixAcc.), mean accuracy (mAcc.), and mean intersection over union (mIoU).

\begin{table}[!t]
	\renewcommand\arraystretch{1.4}
	\begin{center}
		\footnotesize
		\caption{Performance analysis of different task-guided branches in the secondary decoder on NYUDv2 dataset. During the inference, we only obtain the prediction results from the primary decoder for semantic segmentation.}
		\label{tab:decoder}
		\setlength{\tabcolsep}{3mm}{
			\begin{tabular}{lcccc}
				\toprule
				Decoder & PixAcc. & mAcc. & mIoU\\
				\midrule
				Semantic-guided   & 75.9 & 61.6 & 49.0\\
				Depth-guided      & 76.8 & 64.6 & 51.2 \\
				Normal-guided     & \textbf{77.3} & \textbf{64.7} & \textbf{51.5} \\
				Depth-guided+ Normal-guided & 76.8 & 64.0 & 51.0 \\
				\bottomrule
		\end{tabular}}
	\end{center}
\end{table}

\begin{table}[!t]
	\renewcommand\arraystretch{1.4}
	\begin{center}
		\footnotesize
		\caption{Performance analysis for the location of ASPP module (at the end of different Fuse modules) in the primary decoder on NYUDv2 dataset.}
		\label{tab:aspp}
		\setlength{\tabcolsep}{3mm}{
			\begin{tabular}{lcccc}
				\toprule
				ASPP Location & PixAcc. & mAcc. & mIoU\\
				\midrule
				with Fuse2    & 76.2 & 63.5 & 50.1   \\
				with Fuse1    & 76.4 & 64.2 & 50.4   \\
				with Fuse0 (Fig.~\ref{Fig:decoder1})    & \textbf{77.3} & \textbf{64.7} & \textbf{51.5} \\
				\bottomrule
		\end{tabular}}
	\end{center}
\end{table}

\begin{table}[!t]
	\renewcommand\arraystretch{1.4}
	\begin{center}
		\footnotesize
		\caption{Ablation study of the proposed method on NYUDv2 dataset. The Dual-decoder means dual-branch decoder. The FT means fine-tuning stage in our training method.}
		\label{tab:ablation}
		
		\begin{tabular}{lcccc}
			\toprule
			Method & PixAcc. & mAcc. & mIoU \\
			\midrule
			Baseline (Fig.~\ref{fig:intro}(b))+$L_{S}$  & 76.4 & 61.9 & 49.3 \\
			+AFF~\cite{Dai:WACV21}+$L_{S}$              & 77.1 & 63.9 & 50.9 \\
			+SA-Gate~\cite{Chen:ECCV20}+$L_{S}$         & 73.3 & 58.3 & 45.4 \\
			+AMF(SA~\cite{Roy:MICCAI18})+$L_{S}$        & 75.9 & 61.1 & 48.2 \\
			+AMF(CA~\cite{ZhangYL:ECCV18})+$L_{S}$      & 77.2 & 63.3 & 51.0 \\
			+AMF(BAM~\cite{Park:BMVC18})+$L_{S}$        & 76.7 & 64.5 & 50.8 \\
			+AMF(CA)+Dual-decoder+$L$                   & 77.3 & 64.7 & 51.5 \\
			+AMF(CA)+Dual-decoder+FT+$L$                & \textbf{77.5} & \textbf{65.3} & \textbf{52.5} \\
			\bottomrule
		\end{tabular}
	\end{center}
\end{table}

\begin{table*}[!t]
	\renewcommand\arraystretch{1.3}
	\begin{center}
		\caption{Comparison with state-of-the-arts on each category of the NYUDv2 dataset. Percentage (\%) of IoUs are shown for evaluation, with best performance marked in \textbf{bold}.}
		\label{tab:imbalance1}
		\resizebox{\textwidth}{!}{
			\begin{tabular}{lccccccccccccccccccccc}
				\toprule
				Method & \rotatebox{90}{wall} & \rotatebox{90}{floor} & \rotatebox{90}{cabinet} & \rotatebox{90}{bed} & \rotatebox{90}{chair} & \rotatebox{90}{sofa} & \rotatebox{90}{table} & \rotatebox{90}{door} & \rotatebox{90}{window} & \rotatebox{90}{bookshelf} & \rotatebox{90}{picture} & \rotatebox{90}{counter} & \rotatebox{90}{blinds} & \rotatebox{90}{desk} & \rotatebox{90}{shelves} & \rotatebox{90}{curtain} & \rotatebox{90}{dresser} & \rotatebox{90}{pillow} & \rotatebox{90}{mirror} & \rotatebox{90}{floormat}\\
				\midrule
				DeepLab~\cite{Chen:ICL15}            &67.9 &83.0 &53.1 &66.8 &57.8 &57.8 &43.4 &19.4 &45.5 &41.5 &49.3 &58.3 &47.8 &15.5 &7.3  &32.9 &34.3 &40.2 &23.7 &15.0 \\
				FCN~\cite{Shelhamer:TPAMI17}         &69.9 &79.4 &50.3 &66.0 &47.5 &53.2 &32.8 &22.1 &39.0 &36.1 &50.5 &54.2 &45.8 &11.9 &8.6  &32.5 &31.0 &37.5 &22.4 &13.6 \\
				Mutex Constraints~\cite{Deng:ICCV15} &65.6 &79.2 &51.9 &66.7 &41.0 &55.7 &36.5 &20.3 &33.2 &32.6 &44.6 &53.6 &49.1 &10.8 &9.1  &47.6 &27.6 &42.5 &30.2 &32.7 \\
				BI (3000)~\cite{Gadde:ECCV16}        &61.7 &68.1 &45.2 &50.6 &38.9 &40.3 &26.2 &20.9 &36.0 &34.4 &40.8 &31.6 &48.3 &9.3  &7.9  &30.8 &22.9 &19.5 &13.9 &16.1 \\
				LSD-GF~\cite{Cheng:CVPR17}           &78.5 &87.1 &56.6 &70.1 &65.2 &63.9 &46.9 &35.9 &47.1 &\textbf{48.9} &54.3 &66.3 &51.7 &20.6 &13.7 &49.8 &43.2 &50.4 &48.5 &32.2 \\
				STD2P\cite{He:CVPR17}                &72.7 &85.7 &55.4 &73.6 &58.5 &60.1 &42.7 &30.2 &42.1 &41.9 &52.9 &59.7 &46.7 &13.5 &9.4  &40.7 &44.1 &42.0 &34.5 &35.6 \\
				RDFNet~\cite{Lee:ICCV17}             &79.7 &87.0 &60.9 &73.4 &64.6 &65.4 &50.7 &39.9 &\textbf{49.6} &44.9 &61.2 &67.1 &\textbf{63.9} &\textbf{28.6} &14.2 &59.7 &49.0 &49.9 &54.3 &\textbf{39.4} \\
				DeepLab-LFOV~\cite{Chen:TPAMI18}     &70.2 &85.2 &55.3 &68.9 &60.5 &59.8 &44.5 &25.4 &47.8 &42.6 &47.9 &57.7 &52.4 &20.7 &9.1  &36.0 &36.9 &41.4 &32.5 &16.0 \\
				DeepLabV3~\cite{Chen:arXiv17}        &78.8 &83.4 &56.7 &61.9 &57.0 &59.4 &41.3 &39.9 &44.5 &45.1 &60.3 &56.9 &54.9 &22.9 &14.2 &52.4 &40.6 &40.1 &31.3 &30.8 \\
				DCN~\cite{Dai:ICCV17}                &77.0 &83.0 &56.4 &64.7 &57.0 &60.8 &39.9 &35.5 &44.6 &44.7 &59.3 &55.8 &59.9 &20.3 &12.3 &55.9 &51.2 &39.8 &36.2 &34.2 \\
				VCD~\cite{Xiong:CVPR20}              &78.2 &83.7 &57.4 &66.1 &57.2 &60.9 &40.1 &39.5 &45.1 &46.8 &59.4 &58.1 &56.6 &21.9 &16.0 &55.2 &47.0 &42.7 &36.2 &34.3 \\
				ADSD (Ours)                       &\textbf{82.3} &\textbf{87.7} &\textbf{66.5} &\textbf{78.2} &\textbf{66.1} &\textbf{68.3} &\textbf{48.0} &\textbf{44.4} &48.8 &47.1 &\textbf{63.9} &\textbf{71.6} &58.4 &28.5 &\textbf{19.7} &\textbf{66.9} &\textbf{60.0} &\textbf{51.7} &\textbf{58.4} &33.7 \\
				\bottomrule
				\toprule
				Method & \rotatebox{90}{clothes} & \rotatebox{90}{ceiling} & \rotatebox{90}{books} & \rotatebox{90}{fridge} & \rotatebox{90}{tv} & \rotatebox{90}{paper} & \rotatebox{90}{towel} & \rotatebox{90}{shower} & \rotatebox{90}{box} & \rotatebox{90}{board} & \rotatebox{90}{person} & \rotatebox{90}{nightstand} & \rotatebox{90}{toilet} & \rotatebox{90}{sink} & \rotatebox{90}{lamp} & \rotatebox{90}{bathtub} & \rotatebox{90}{bag} & \rotatebox{90}{ot. struct.} & \rotatebox{90}{ot. furn.} & \rotatebox{90}{ot. props.}\\
				\midrule
				DeepLab~\cite{Chen:ICL15}             &20.2 &55.1 &22.1 &30.6 &49.4 &21.8 &32.1 &6.4  &5.8  &14.8 &55.3 &37.7 &57.9 &47.7 &40.0 &44.7 &6.6  &18.0 &12.9 &33.8\\
				FCN~\cite{Shelhamer:TPAMI17}          &18.3 &59.1 &27.3 &27.0 &41.9 &15.9 &26.1 &14.1 &6.5  &12.9 &57.6 &30.1 &61.3 &44.8 &32.1 &39.2 &4.8  &15.2 &7.7  &30.0 \\
				Mutex Constraints~\cite{Deng:ICCV15}  &12.6 &56.7 &8.9  &21.6 &19.2 &28.0 &28.6 &22.9 &1.6  &1.0  &9.6  &30.6 &48.4 &41.8 &28.1 &27.6 &0    &9.8  &7.6  &24.5 \\
				BI (3000)~\cite{Gadde:ECCV16}         &13.7 &42.5 &21.3 &16.6 &30.9 &14.9 &23.3 &17.8 &3.3  &9.9  &44.7 &15.8 &53.8 &32.1 &22.8 &19.0 &0.1  &12.3 &5.3  &23.2 \\
				LSD-GF~\cite{Cheng:CVPR17}            &24.7 &62.0 &34.2 &45.3 &53.4 &27.7 &\textbf{42.6} &23.9 &11.2 &58.8 &53.2 &54.1 &\textbf{80.4} &59.2 &45.5 &52.6 &15.9 &12.7 &16.4 &29.3 \\
				STD2P\cite{He:CVPR17}                 &22.2 &55.9 &29.8 &41.7 &52.5 &21.1 &34.4 &15.5 &7.8  &29.2 &60.7 &42.2 &62.7 &47.4 &38.6 &28.5 &7.3  &18.8 &5.1  &31.4 \\
				RDFNet~\cite{Lee:ICCV17}              &\textbf{26.9} &69.1 &\textbf{35.0} &\textbf{58.9} &63.8 &\textbf{34.1} &41.6 &38.5 &11.6 &54.0 &\textbf{80.0} &45.3 &65.7 &62.1 &\textbf{47.1} &57.3 &\textbf{19.1} &30.7 &20.6 &39.0 \\
				DeepLab-LFOV~\cite{Chen:TPAMI18}      &17.8 &58.4 &20.5 &45.1 &48.0 &21.0 &41.5 &9.4  &8.0  &14.3 &67.0 &41.8 &69.7 &46.8 &40.1 &45.1 &2.1  &20.7 &12.4 &33.5 \\
				DeepLabV3~\cite{Chen:arXiv17}         &20.7 &69.8 &30.3 &42.8 &52.5 &27.7 &33.2 &24.5 &13.6 &68.9 &73.3 &37.7 &65.1 &51.3 &39.2 &36.4 &12.5 &27.7 &15.2 &36.6 \\
				DCN~\cite{Dai:ICCV17}                 &22.3 &63.3 &26.9 &52.8 &58.7 &29.9 &39.8 &40.4 &14.9 &65.3 &76.2 &39.9 &67.1 &50.3 &38.7 &40.1 &7.3 &26.7 &16.5 &36.9 \\
				VCD~\cite{Xiong:CVPR20}               &22.2 &67.0 &30.0 &50.9 &57.0 &30.7 &36.7 &40.6 &\textbf{15.6} &72.6 &77.5 &41.2 &69.1 &51.8 &43.0 &39.4 &9.5 &27.7 &18.3 &37.0 \\
				ADSD (Ours)                        &24.0 &\textbf{76.0} &32.9 &57.8 &\textbf{70.8} &28.6 &40.3 &\textbf{48.2} &12.1 &\textbf{78.3} &67.3 &\textbf{57.1} &77.9 &\textbf{63.2} &46.5 &\textbf{62.2} &9.6  &\textbf{33.4} &\textbf{22.2} &\textbf{39.6} \\
				\bottomrule
		\end{tabular}}
	\end{center}
\end{table*}

\begin{table*}[!t]
	\renewcommand\arraystretch{1.3}
	\begin{center}
		\caption{Comparison with state-of-the-arts on each category of the SUN-RGBD dataset. Percentage (\%) of IoUs are shown for evaluation, with best performance marked in \textbf{bold}.}
		\label{tab:imbalance2}
		\resizebox{\textwidth}{!}{
			\begin{tabular}{lccccccccccccccccccccc}
				\toprule
				Method & \rotatebox{90}{wall} & \rotatebox{90}{floor} & \rotatebox{90}{cabinet} & \rotatebox{90}{bed} & \rotatebox{90}{chair} & \rotatebox{90}{sofa} & \rotatebox{90}{table} & \rotatebox{90}{door} & \rotatebox{90}{window} & \rotatebox{90}{bookshelf} & \rotatebox{90}{picture} & \rotatebox{90}{counter} & \rotatebox{90}{blinds} & \rotatebox{90}{desk} & \rotatebox{90}{shelves} & \rotatebox{90}{curtain} & \rotatebox{90}{dresser} & \rotatebox{90}{pillow} & \rotatebox{90}{mirror} \\
				\midrule
				Song \emph{et al.}~\cite{Song:CVPR15}  &36.4 &45.8 &15.4 &23.3 &19.9 &11.6 &19.3 &6.0 &7.9 &12.8 &3.6 &5.2 &2.2 &7.0 &1.7 &4.4 &5.4 &3.1 &5.6\\
				Liu \emph{et al.}~\cite{Liu:TPAMI11}   &37.8 &48.3 &17.2 &23.6 &20.8 &12.1 &20.9 &6.8 &9.0 &13.1 &4.4 &6.2 &2.4 &6.8 &1.0 &7.8 &4.8 &3.2 &6.4\\
				Ren \emph{et al.}~\cite{Ren:CVPR12}    &43.2 &78.6 &26.2 &42.5 &33.2 &40.6 &34.3 &33.2 &43.6 &23.1 &57.2 &31.8 &42.3 &12.1 &18.4 &59.1 &31.4 &49.5 &24.8\\
				DeconvNet~\cite{Cheng:CVPR17}          &90.4 &92.7 &57.7 &75.9 &83.0 &61.2 &64.2 &43.0 &64.7 &42.3 &59.8 &42.5 &48.3 &29.5 &17.5 &64.9 &54.0 &61.7 &51.3\\
				LSD-GF~\cite{Cheng:CVPR17}      &91.9 &94.7 &61.6 &82.2 &87.5 &62.8 &68.3 &47.9 &68.0 &48.4 &69.1 &49.4 &51.3 &\textbf{35.0} &\textbf{24.0} &68.7 &60.5 &\textbf{66.5} &57.6\\
				ADSD (Ours)                       &\textbf{92.1} &\textbf{96.0} &\textbf{70.9} &\textbf{84.0} &\textbf{86.7} &\textbf{74.5} &\textbf{72.5} &\textbf{58.5} &\textbf{70.4} &\textbf{51.7} &\textbf{71.8} &\textbf{57.0} &\textbf{54.3} &29.6 &21.6 &\textbf{78.1} &\textbf{67.2} &64.9 &\textbf{64.0} \\
				\bottomrule
				\toprule
				Method & \rotatebox{90}{floormat} & \rotatebox{90}{clothes} & \rotatebox{90}{ceiling} & \rotatebox{90}{books} & \rotatebox{90}{fridge} & \rotatebox{90}{tv} & \rotatebox{90}{paper} & \rotatebox{90}{towel} & \rotatebox{90}{shower} & \rotatebox{90}{box} & \rotatebox{90}{board} & \rotatebox{90}{person} & \rotatebox{90}{nightstand} & \rotatebox{90}{toilet} & \rotatebox{90}{sink} & \rotatebox{90}{lamp} & \rotatebox{90}{bathtub} & \rotatebox{90}{bag} &\rotatebox{90}{mACC.}\\
				\midrule
				Song \emph{et al.}~\cite{Song:CVPR15}&0 &1.4 &35.8 &6.1 &9.5 &0.7 &1.4 &0.2 &0.0 &0.6 &7.6 &0.7 &1.7 &12.0 &15.2 &0.9 &1.1 &0.6 &9.0\\
				Liu \emph{et al.}~\cite{Liu:TPAMI11} &0 &1.6 &49.2 &8.7 &10.1 &0.6 &1.4 &0.2 &0.0 &0.8 &8.6 &0.8 &1.8 &14.9 &16.8 &1.2 &1.1 &1.3 &10.1\\
				Ren \emph{et al.}~\cite{Ren:CVPR12}  &\textbf{5.6} &27.0 &84.5 &35.7 &24.2 &36.5 &26.8 &19.2 &9.0 &11.7 &51.4 &35.7 &25.0 &64.1 &53.0 &44.2 &47.0 &18.6 &36.3\\
				DeconvNet~\cite{Cheng:CVPR17}        &0.4 &39.8 &78.3 &55.0 &43.9 &59.6 &29.4 &45.2 &1.5 &35.9 &47.7 &45.3 &36.0 &77.6 &66.6 &51.2 &66.1 &35.8 &51.9
				\\
				LSD-GF~\cite{Cheng:CVPR17}      &0 &44.7 &\textbf{88.8} &\textbf{61.5} &51.4 &\textbf{71.7} &37.3 &\textbf{51.4} &2.9 &\textbf{46.0} &54.2 &49.1 &44.6 &82.2 &74.2 &\textbf{64.7} &77.0 &\textbf{47.6} &58.0\\
				ADSD (Ours)                         &0 &\textbf{55.2} &87.6 &59.6 &\textbf{66.9} &68.6 &\textbf{43.4} &49.8 &\textbf{29.5} &45.9 &\textbf{64.6} &\textbf{62.6} &\textbf{46.8} &\textbf{88.1} &\textbf{76.4} &62.8 &\textbf{84.2} &42.6 &\textbf{62.1} \\
				\bottomrule
		\end{tabular}}
	\end{center}
\end{table*}

\begin{figure*}[!t]
	\centering
	\includegraphics[width=6.3in]{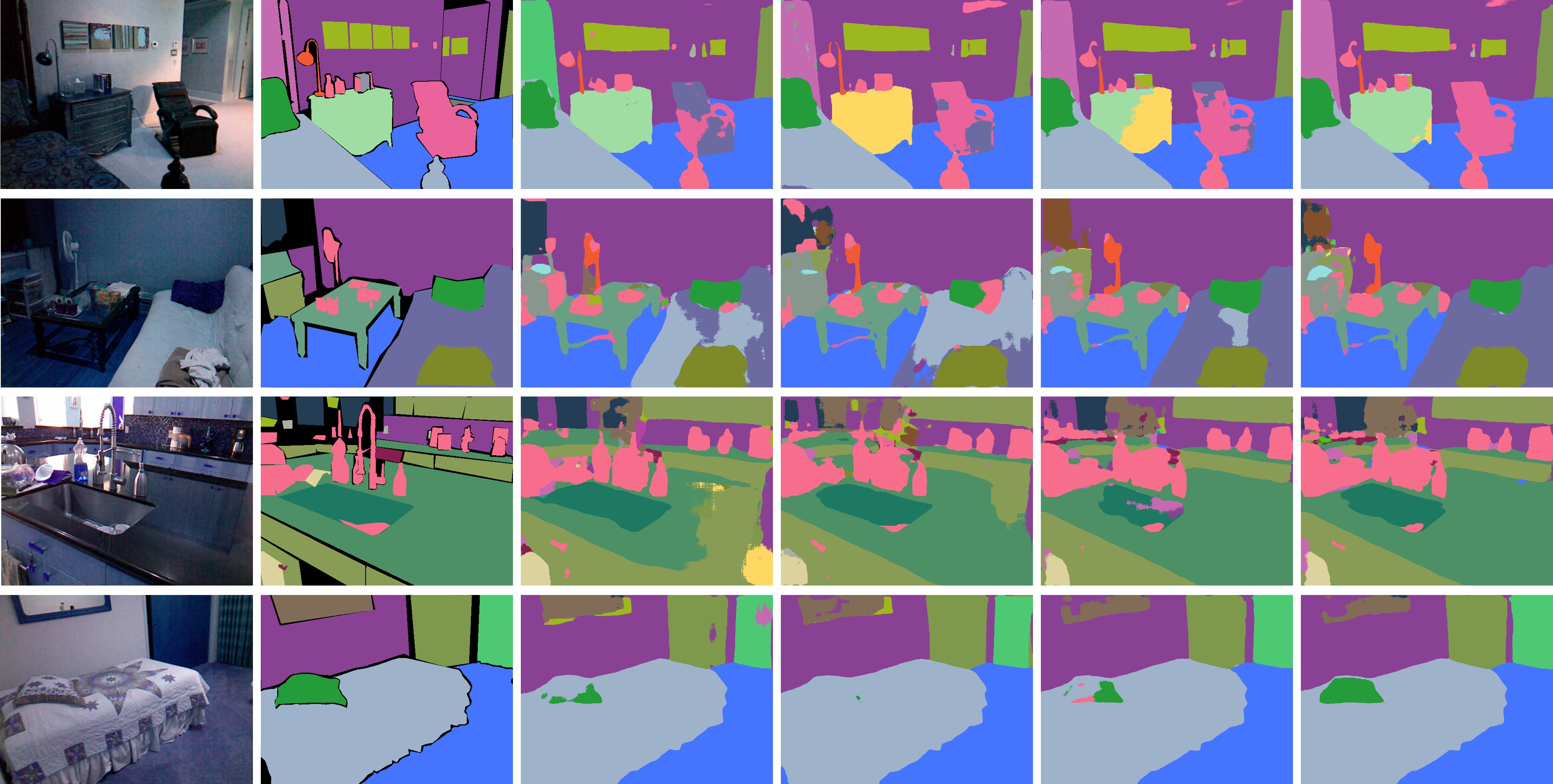}   		
	\caption{The visual results of ablation analysis on NYUDv2 dataset. From left to right, we show the inputs, ground-truths, the results of baseline, with AMF, with AMF and dual-branch decoder, and our method, respectively.}
	\label{fig:ablation}
\end{figure*}

\subsection{Ablation study}\label{sec:ablation}
To discover the functionality of each component in our method, we conduct an ablation study on the NYUDv2 dataset. Taking the network consisting of a two-stream encoder and a simple decoder (see Fig.~\ref{fig:intro}(b)) as a baseline. In the encoder, the combination operation of multi-modal features is the element-wise summation (like FuseNet~\cite{Hazirbas:ACCV17}). We evaluate the effectiveness of our dual-branch decoder by choosing different task-guided branches and changing the location of AASP module. The results can be found in Table~\ref{tab:decoder} and Table~\ref{tab:aspp}. For task-guided branches of the secondary decoder, the normal-guided performs better than the depth-guided, semantic-guided and the combination of depth-guided and semantic-guided. For the location of ASPP (at the end of different Fuse modules) in the primary decoder, the ASPP with Fuse0 (in Fig.~\ref{Fig:decoder1}) performs better than the ASPP with Fuse1 and Fuse2 modules.

The results of ablation analysis are shown in Table~\ref{tab:ablation}. For the multi-model feature fusion solutions (Fig.~\ref{Fig:Attention}), our AMF with channel attention (CA) performs better than the element-wise summation (\textit{i.e.} baseline), separation-and-aggregation gate~\cite{Chen:ECCV20}, attentional feature fusion (AFF)~\cite{Dai:WACV21}, bottleneck attention module (BAM)~\cite{Park:BMVC18}, and spatial attention (SA). We owe this to the representation of channel attention that ignores less important channels of fused features and emphasizes the important ones. The results demonstrate that our AMF can significantly improve the performance of semantic segmentation. This observation also clarifies that incorporating depth information can greatly improve the performance, which reveals the effectiveness of reasoning color and geometry information together. By introducing the dual-branch decoder, the performance is further improved. The final fine-tuning (FT) stage of training strategy (see details in Section~\ref{sec:decoder}) for our model gives another rise in the performance.

\begin{table*}[!t]
	\renewcommand\arraystretch{1.3}
	\begin{center}
		\footnotesize
		\caption{Comparison with state-of-the-arts on NYUDv2 test set in 40-class and SUN-RGBD test set in 37-class. Percentage (\%) of PixAcc., mAcc., and mIoU are shown for evaluation. In category, the `AC', `ED', and `MT' denote atrous/dilated convolution, encoder-decoder, and encoder-decoder for multi-task, respectively. In scale, the `S', and `M' denote single-scale inference strategy and multi-scale inference strategy, respectively.}
		\label{tab:sota}
		\setlength{\tabcolsep}{3mm}{
			\begin{tabular}{lccccccccccc}
				\toprule
				\multirow{2}{*}{{\begin{tabular}[l]{@{}c@{}} Method \end{tabular}}}   & \multirow{2}{*}{{\begin{tabular}[c]{@{}c@{}} Category \end{tabular}}} & \multirow{2}{*}{{\begin{tabular}[c]{@{}c@{}} Data\end{tabular}}}      &
				\multirow{2}{*}{{\begin{tabular}[c]{@{}c@{}} Backbone\end{tabular}}}  &
				\multirow{2}{*}{{\begin{tabular}[c]{@{}c@{}} Scale \end{tabular}}}    &
				\multicolumn{3}{c}{NYUDv2 (40-class)} &
				\multicolumn{3}{c}{SUN-RGBD (37-class)}\\ \cline{6-11}
				& & & & & PixAcc. & mAcc. & mIoU & PixAcc. & mAcc. & mIoU \\
				\midrule
				DeepLab~\cite{Chen:ICL15}        & AC & RGBD & VGG-16     & M & 68.7 & 46.9 & 36.8 & --   & --   & --   \\
				BI (3000)~\cite{Gadde:ECCV16}    & AC & RGBD & VGG-16     & S & 58.9 & 39.3 & 27.7 & --   & --   & --   \\
				CFN\cite{Lin:ICCV17}             & AC & RGBD & VGG-16     & M & --   & --   & 41.7 & --   & --   & 42.5 \\
				3DGNN~\cite{Qi:ICCV17}           & AC & RGBD & VGG-16     & M & --   & 55.7 & 43.1 & --   & 54.6 & 42.3 \\
				DeepLab-LFOV~\cite{Chen:TPAMI18} & AC & RGBD & VGG-16     & M & 70.3 & 49.6 & 39.4 & 71.9 & 42.2 & 32.1 \\
				D-CNN~\cite{Wang:ECCV18}         & AC & RGBD & VGG-16     & S & --   & 56.3 & 43.9 & --   & 53.5 & 42.0 \\
				RefineNet~\cite{Lin:TPAMI20}     & AC & RGB  & ResNet-152 & M & 74.4 & 59.6 & 47.6 & 81.1 & 57.7 & 47.0 \\
				\midrule
				DeconvNet~\cite{Noh:ICCV15}      & ED & RGB  & VGG-16     & S & --   & --   & --   & 66.1 & 32.3 & 22.6 \\
				FCN~\cite{Shelhamer:TPAMI17}     & ED & RGBD & VGG-16     & S & 65.4 & 46.1 & 34.0 & 68.2 & 38.4 & 27.4 \\
				SegNet~\cite{Badri:TPAMI17}      & ED & RGB  & VGG-16     & S & --   & --   & --   & 72.6 & 44.8 & 31.8 \\
				B-SegNet~\cite{Kendall:BMVC17}   & ED & RGB  & VGG-16     & S & 68.0 & 45.8 & 32.4 & 71.2 & 45.9 & 30.7 \\
				FuseNet~\cite{Hazirbas:ACCV17}   & ED & RGBD & VGG-16     & S & --   & --   & --   & 76.3 & 48.3 & 37.3 \\
				LSD-GF~\cite{Cheng:CVPR17}       & ED & RGBD & VGG-16     & S & 71.9 & 60.7 & 45.9 & --   & 58.0 & --   \\
				RDFNet-152~\cite{Lee:ICCV17}     & ED & RGB  & ResNet-152 & M & 76.0 & 62.8 & 50.1 & 81.5 & 60.1 & 47.7 \\
				RedNet~\cite{Jiang:arXiv18}      & ED & RGBD & ResNet-50  & S & --   & 62.6 & 47.2 & 81.3 & 60.3 & 47.8 \\
				ACNet~\cite{Hu:ICIP19}           & ED & RGBD & ResNet-50  & S & --   & 63.1 & 48.3 & --   & 60.3 & 48.1 \\
				CANet~\cite{Zhou:ACCV20}         & ED & RGBD & ResNet-101 & S & 76.6 & 63.8 & 51.2 & 82.5 & 60.5 & 49.3 \\
				SGNet~\cite{Chen:arXiv20}        & ED & RGBD & ResNet-101 & S & 76.4 & 62.7 & 50.3 & 81.0 & 59.6 & 47.1 \\
				Malleable 2.5D~\cite{Xing:ECCV20}& ED & RGBD & ResNet-101 & M & 76.9 & --   & 50.9 & --   & --   & --   \\
				SA-Gate~\cite{Chen:ECCV20}       & ED & RGBD & ResNet-101 & M & --   & --   & 52.4 & -- & --   & 49.4 \\
				ESANet~\cite{Seichter:arXiv20}   & ED & RGBD & ResNet-50  & S & --   & --   & 50.5 & --   & --   & 48.3   \\
				\midrule
				MS CNN~\cite{Eigen:ICCV15}       & MT & RGB  & VGG-16     & S & 65.6 & 45.1 & 34.1 & --   & --   & --   \\
				PU-Loop~\cite{Kong:CVPR18}       & MT & RGB  & ResNet-50  & S & 72.1 & --   & 44.5 & 80.3 & --   & 45.1 \\
				TRL~\cite{Zhang:ECCV18}          & MT & RGB  & ResNet-50  & S & 76.2 & 56.3 & 46.4 & 83.6 & 58.2 & 49.6 \\
				PAD-Net~\cite{Xu:CVPR18}         & MT & RGB  & ResNet-50  & S & 75.2 & 62.3 & 50.2 & --   & --   & --   \\
				RTJ-AA~\cite{Nekrasov:ICRA19}    & MT & RGB & MobileNetV2 & S & --   & --   & 42.0 & --   & --   & --   \\
				PAP~\cite{Zhang:CVPR19}          & MT & RGB  & ResNet-50  & S & 76.2 & 62.5 & 50.4 & 83.8 & 58.4 & 50.5 \\
				PSD~\cite{Zhou:CVPR20}           & MT & RGB  & ResNet-50  & S & 77.0 & 58.6 & 51.0 & \textbf{84.0} & 57.3 & \textbf{50.6} \\
				MTI-Net~\cite{Vandenhende:ECCV20}& MT & RGB  & HRNet48-V2 & S & 75.3 & 62.9 & 49.0 & --   & --   & --   \\
				\midrule
				ADSD (Ours)                      & ED & RGBD & ResNet-50  & S & \textbf{77.5} & \textbf{65.3} & \textbf{52.5} & 81.8 & \textbf{62.1} & 49.6\\
				\bottomrule
		\end{tabular}}
	\end{center}
\end{table*}

\begin{figure}[!t]
	\begin{center}
		\includegraphics[width=3.1in]{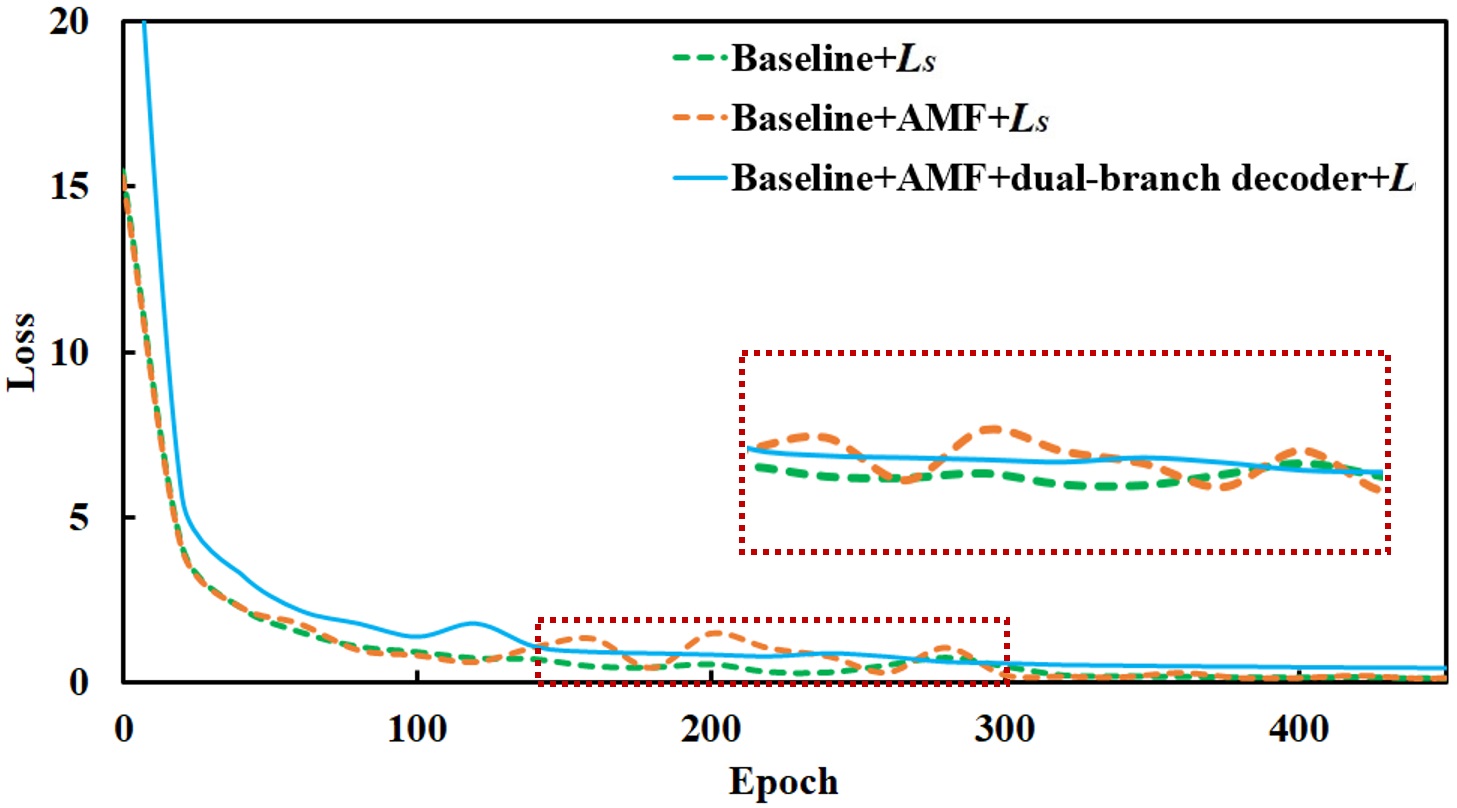}
	\end{center}
	\caption{Statistics of loss values during a training procedure on NYUDv2 dataset. Our dual-branch decoder can reduce the adverse effects on imbalance between encoder and decoder to facilitate the convergence of training.}
	\label{fig:loss}
\end{figure}

We show qualitative results of our method on NYUDv2 dataset for semantic segmentation in Fig.~\ref{fig:ablation}. For comparison, we also include the visual results of baseline, with the proposed AMF, with the AMF and dual-branch decoder, with the AMF, dual-branch decoder and fine-tuning (FT) stage (our method). The results show that the geometry information is well distilled by our AMF, which can distinguish the objects with similar color. Moreover, when incorporating the dual-branch decoder, our network can recover more context information and more accurate object masks.
As mentioned before, we argue that the training is going faster and more robust along with our dual-branch decoder. This decoder can also deal with the imbalance problem caused by the phenomenon that the encoder uses multi-modal information, while the decoder dose not. To verify this statement, we report the loss values of our method and the methods without a dual-branch decoder during training. As shown in Fig.~\ref{fig:loss}, we observe that the multi-loss $L$ of our method is rapidly reduced only after 150 epochs from the beginning. However, the loss of the methods without a dual-branch decoder waves violently due to the imbalance of encoder and decoder. We also find that the dual-branch decoder can facilitate the convergence of training, which is capable of reducing the adverse effects on this imbalance.

To evaluate the performance of our model on the imbalanced distributed data, we also show the results on each category, as shown in Table~\ref{tab:imbalance1}. Clearly, our method performs better than other methods in most categories. Specially, our method still achieves a relatively higher IoU on some “hard” categories such as bed, curtain, dresser, shower, and board, etc. Following previous methods~\cite{Song:CVPR15,Liu:TPAMI11,Ren:CVPR12,Cheng:CVPR17}, we also report the mACC. of our method on SUN-RGBD dataset. As shown in Table~\ref{tab:imbalance2}, we achieve 62.1\% mean accuracy with 4.1\% improvement over the recent method~\cite{Cheng:CVPR17}. Specifically, we yield performance gains over 26 classes, which demonstrates the effectiveness of the proposed approach.
We owe the robustness among almost all the categories to the effectively learned multi-modal cues in the encoder, and the cross-modal information in the decoder. Note that our method achieves unsatisfactory performance on some categories (\textit{e.g.} blinds, person, bag) of NYUDv2 dataset, which may due to our joint reasoning on color and geometric cues, as the depth may vary greatly compared with the corresponding color appearance in different scenes.

\subsection{Compared with State-of-the-arts}\label{sec:sota}
\noindent
\textbf{NYUDv2.}
The comparison results on the NYUDv2 dataset with 40-category are shown in Table~\ref{tab:sota}. We use ResNet-50 and single-scale inference strategy for a fair comparison. Following our training method mentioned in Section~\ref{sec:decoder}, we use a normal-guided branch decoder at the pre-training stage and the semantic-guided branch decoder at the fine-tuning stage. Our method can still achieve 77.5\% PixAcc., 65.3\% mAcc., and 52.5\% mIoU, which is better than the state-of-the-art methods. Specifically, we can find that utilizing depth and normal as extra supervision could make network more robust than general RGBD methods that take both RGB and depth as inputs. Besides, it can be observed that the methods try to use atrous/dilated convolution or gate fusion to extract complementary feature, which are more implicit than our model in selecting valid feature from complementary information.

\noindent
\textbf{SUN-RGBD.}
We also compare our method with the state-of-the-arts on the large-scale SUN-RGBD dataset. Due to the lacking of surface normal ground-truths, we use a depth-guided branch decoder at the pre-training stage and the semantic-guided branch decoder at the fine-tuning stage. As summarized in Table~\ref{tab:sota}, our ADSD achieves 81.8\% PixAcc., 62.1\% mAcc., and 49.6\% mIoU, which is the best results on mAcc. in comparison with the pervious methods. Moreover, our method obtains the superior performance than the approaches based on atrous/dilated convolution and encoder-decoder network, suggesting its superiority and high performance for RGBD semantic segmentation. We can observe that our proposed ADSD is slightly weaker than multi-task learning based methods such as PAP~\cite{Zhang:CVPR19} and PSD~\cite{Zhou:CVPR20} on both PixAcc. and mIoU metrics. The main reason is that we perform the dual-supervised decoder with a depth-guided branch at the pre-training stage on SUN-RGBD dataset. Note that there are many low-quality depth maps in SUN-RGBD dataset caused by the capture device~\cite{Song:CVPR15,Lee:ICCV17}, which may affect the auxiliary utility from the depth.
More details of qualitative results are shown in the supplementary material. 

\section{Conclusions}\label{sec:conclusions}
In this paper, we have proposed a novel encoder-decoder framework for RGBD semantic segmentation, which can take full advantage of the complementary information across modalities. The color and depth data were jointly reasoned by forming a two-stream encoder. The multi-level paired complementary cues can be processed by our proposed AMF in the encoder. We then introduced a dual-branch decoder to effectively leverage the correlation and complementation of different tasks. In the decoder, the primary branch was used to incorporate multi-scale context by the ASPP with pyramid supervision. In addition, it was further supervised by another task-branch like normal estimation to improve the performance of segmentation and training convergence speed. Experiments on NYUDv2 and SUN-RGBD datasets demonstrated the superiority of our method compared with the previous approaches on RGBD semantic segmentation. In the future, we will generalize our method on more vision tasks and improve its efficiency.


{\small
\bibliographystyle{cvm}
\bibliography{egbib.bib}
}

\end{document}